\documentclass[10pt]{paper}
\usepackage[utf8]{inputenc}
\usepackage[version=4]{mhchem}
\usepackage{graphicx}
\usepackage{amsmath}
\usepackage{float}
\usepackage{comment}
\usepackage[justification=centerfirst]{subfig}
\usepackage{cite}
\usepackage[a4paper, margin=2cm]{geometry}

\usepackage{psfrag}
\usepackage{caption}
\usepackage[dvipsnames]{xcolor}
\usepackage{multirow}
\usepackage{natbib}
\usepackage{amssymb}
\usepackage{yfonts}
\usepackage{tikz}
 
\newcommand{\one}{($i$) }
\newcommand{\two}{($ii$) }
\newcommand{\three}{($iii$) }

\title{A thermodynamically consistent chemical spiking neuron capable of autonomous Hebbian learning}
\author{Jakub Fil and Dominique Chu
\\
CEMS, School of Computing, University of Kent, CT2 7NF, Canterbury, UK}
\date{\today}
\begin{document}

\maketitle
\begin{abstract}
We propose a fully autonomous, thermodynamically consistent set of chemical reactions that implements a spiking neuron. This chemical neuron is able to learn input patterns in a Hebbian fashion. The system is scalable to arbitrarily many input channels. We demonstrate its performance in learning  frequency biases in the input as well as correlations between different input channels.   Efficient computation of time-correlations requires a highly non-linear activation function. The resource requirements of a non-linear activation function are discussed.  In addition to the thermodynamically consistent  model of the CN, we also propose  a biologically plausible version that could be engineered in a synthetic biology context.
\end{abstract}
\section{Introduction}

Beside electronic devices there are a number of substrates that can be used to realise computations \citep{adamandeve}. A particularly important class of what is sometimes called ``unconventional computers'' are chemical systems and especially biochemical systems \citep{amos}. These underpin the remarkable abilities of unicellular organisms to adapt to changes in their environment in the absence of a nervous system. Examples of biochemical information processors include sensing  \citep{government1, government2, alon2019introduction}, chemotaxis  \citep{Yi_2000, Hoffer_2001} or diauxic growth \citep{my1, my2, my3}.  Understanding the function of biochemical systems has so far been the domain of molecular biology.  With the advent of synthetic biology it is now possible  to build custom-made  biochemical systems and hence  to engineer wet-ware molecular computers. There are a number of possible applications of such computers. Particularly promising areas include personalised precision medicine, especially  targeted drug delivery but also environmental clean-up or sensing.  Before any such computers can be built, it is first necessary to understand how to program them.  
\par
Here we propose a chemical reaction network --- henceforth referred to as the {\em chemical neuron} (CN) ---  that behaves  like  an artificial   spiking neuron (SN).  SNs  are a type of  learning machine that is used widely in Artificial Intelligence as a component of neural networks \citep{deep_snn, afshar_snn}. It is now also well established that a single SN has significant  learning capabilities \citep{Fil_2020, Gutig2016}. In computer science they  have been applied to a number of real-world tasks including principal component analysis \citep{oja_1982}, recognition of handwriting \citep{diehl_digits} or classification of  fighter-planes \citep{van_schaik_planes}. There are many different models of  SNs in the literature. Commonly a SN  has  an internal state which is typically represented by a positive number.  The input state may decay, that means that it  reduces over time with some rate.  The input can also increase when the SN receives a stimulus (an input {\em spike}) via one of its $N$ input channels. Importantly, these input channels are weighted such that  the increase of the internal state variable is different for the different input channels. The higher the weight of a channel, the stronger a stimulus will affect the internal state. 
\par
A central problem of research in neural networks is to find  ways to to determine the weights of  a neuron such that it performs a particular task. The process of finding these weights is  usually referred to as the ``training'' or alternatively as the learning phase.   A common method  to train SNs is {\em Hebbian learning}. The basic idea of Hebbian learning is that the weights relating the input to the output are plastic,  i.e. they change over time. In Hebbian learning, the weights will be strengthen by some amount if the input to the neuron coincides with the neuron generating an output event. 
A well-known special case of Hebbian learning is associative learning. Assume that $A_1$ and $A_2$ are both inputs to B. Assume further that the connection between $A_1$ is sufficiently strong such that its firing can trigger a firing event of  B. If  $A_2$ fires usually  at around the same time as A$_1$, then its weights will be strengthened. Eventually, the rates of the second channel will have adjusted sufficiently such that  firing of $A_2$ alone  will be sufficient to trigger a response on its own. 
\par
There have been a number of previous attempts to design chemical implementations of learning machines.   Amongst the earliest is perhaps work in the 1980s by Okamoto and collaborators  \citep{Okamoto_1988} who showed that certain biochemical systems implement the McCulloch-Pitts neuronic equations.    A chemical neuron  has been implemented by  \citep{Hjelmfelt_1991}; however, this neuron could only detect stable concentrations of input molecules and no learning was shown.   Fernando and coworkers  \citep{Fernando2009}  showed  that a relatively simple set of molecular interactions can implement associative learning. Their model is fully autonomous, but it is also inflexible.  Association is learned after  just a single coincidence, and hence not fit to detect  statistical correlations robustly.   Moreover, the system cannot forget the association between the inputs. \citep{mcgregor_cn} improved on that with systems that were found by evolutionary processes. A biochemically plausible implementation of the associative learning based on genetic regulatory networks  was proposed by \citep{macia1, macia2,solebiochemimp}. They imagine their system not to be within a single cell, but distributed across multiple compartments. This has the advantage that is  reduces crosstalk between molecular species and allows more complicated networks to be implemented.  Another model close to biology is by Nesbeth  \citep{assmodel2}.  A  perceptron formulated in an artificial chemistry was     proposed  by  Banda {\em et al.} \citep{chemperceptron}; later this was extended to a  fully fledged feed-forward neural network  \citep{chemperceptron2} which could solve the XOR problem.  Their model still requires regular intervention by outside operators. Besides these simulation studies,   there have also been attempts to implement learning {\em in vivo}  \citep{assmodel,assmodel2,biochemimp,forcement}.   
\par
The main contribution of this article is a design  for a novel  chemical neuron (CN), that is a family of  molecular reaction networks.  The CN implements a  fully autonomous  spiking neuron capable of  full Hebbian learning. A tuneable parameter of the design of the CN is the non-linearity of the {\em activation function}. This activation function determines the probability that the CN generates  a learning signal, depending on the internal state. 
\par
We demonstrate that the CN is capable of Hebbian learning, in the  sense that its internal molecular abundances will reflect statistical biases of its inputs.  More specifically, via each its input channels the CN can accept   {\em boli}  of some chemical species.  Over time these boli will lead to a reconfiguration of the  steady state abundances of the CN. These abundances will then be reflective of  statistical biases of the input boli.  In particular we consider two types of biases: \one Frequency biases (FB), where the average waiting time between two boli is different for different channels. Here the input times are independent but not identically distributed random variables.  \two {\em Time Correlations} (TC), where the frequency of boli may be the same for each input channel, but inputs for one channel tend to happen before or after the inputs of one or more other channels. Here, input boli are identically distributed, but not independently drawn from one another. We find that the CN can identify both types of biases. However, we also show that the FB task is trivial for chemical systems, whereas the TC task can only be solved when the non-linearity is high. We find that a high non-linearity comes with minimal resource requirements. 
\par
In this contribution, we will give two versions of the CN. The first (basic) version will be a minimal set of reactions. It is thermodynamically consistent in that it consists only of micro-reversible reactions with mass-action kinetics. This basic version cannot be realised easily though. Therefore, we shall propose a second version of the model, which is not thermodynamically explicit, but biologically plausible and can likely be engineered in a synthetic biology framework.  
\section{Results}
\begin{figure}
\psfrag{B}{$E$}
\includegraphics[width=\textwidth]{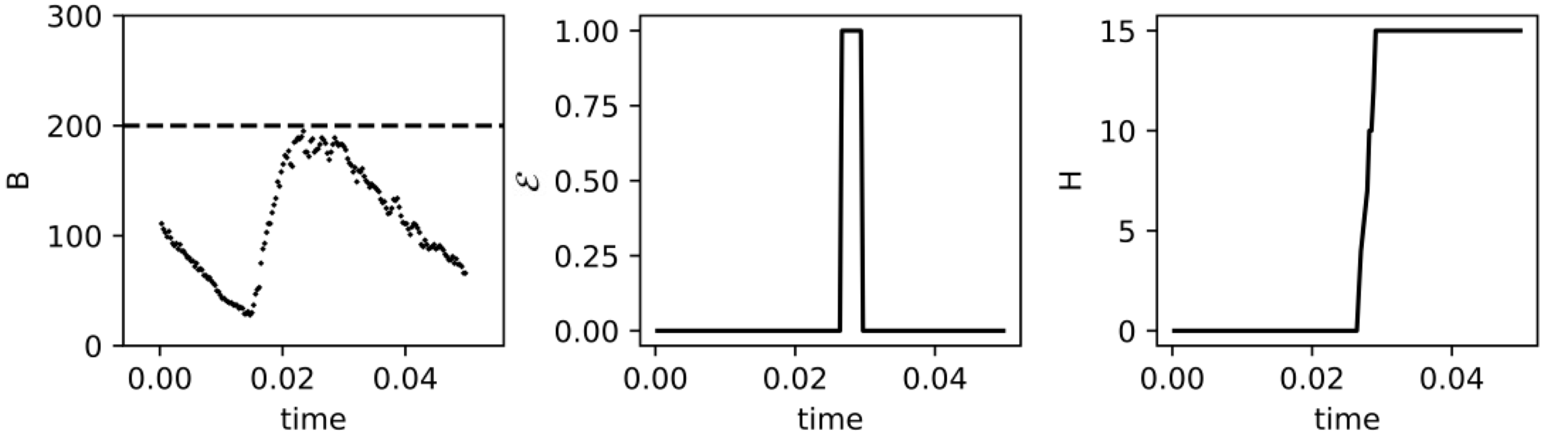}
\caption{
Example simulation showing the core of the CN dynamics. The graphs show the internal state $B$, the learning signal $\mathcal E$ and the weight $H$ for a single channel. We assume a bolus provided at time $t=0.015$. This causes the internal state to go up and reach the threshold. A learning signal is triggered at around $t=0.03$ and consequently the weight is increase by (in this case) 15 molecules of $H$. }
\label{example_CRN_small}
\end{figure}
\subsection{Model Description}
\begin{table}
\centering
\begin{tabular}{|l|l|} 
\hline
Input  & \ce{A_n <=>[k_{AB}][k_{BA}] B}  \\ 
\hline
\multirow{2}{*}{Activation function }
& \ce{B + E_{i} <=>[k^{+}][k^{-}] E_{i+1}}, \qquad $i< m-1$ \\ \cline{2-2}
& \ce{B + E_{m-1} <=>[k^{+}][k^{-}_{last}] $\mathcal{E}$}   \\ 
\hline
\multirow{2}{*}{Learning}
& \ce{A_n + $\mathcal{E}$ <=>[k_{AE}][k_{EA}] $A\mathcal{E}_{n}$ <=>[k_{EH}][k_{HE}] H_n + $\mathcal{E}$}   \\ \cline{2-2}
& \ce{A_n + H_n <=>[k_{AH}][k_{HA}] AH_{n} <=>[k_{HB}][k_{BH}] B + H_n}   \\ 
\hline
\multirow{2}{*}{Leak}
& \ce{H_n ->[k_{H$\varnothing$}] $\varnothing$}   \\ \cline{2-2}
& \ce{B ->[k_{B$\varnothing$}] $\varnothing$}   \\ 
\hline
\end{tabular}
\caption{List of chemical reactions in a single CN.}
\label{reaction_table}
\end{table}
%
\paragraph{Overview}
We implement the CN as a set of micro-reversible elementary chemical reactions obeying mass-acting kinetics. Micro-reversibility makes the model thermodynamically consistent. See table \ref{reaction_table} for the list of  reactions.  The system is best understood by thinking of  the molecular species $A_i$ as the  input to the system via input channel $i$. The weight equivalent of the $i$-th input channel of the  CN are the abundances of the species $H_i$. The species $\mathcal E$ is the activated form of $E$ and takes the dual role as the learning signal and the output of the CN.  The internal state of the CN is represented by the  abundance of the molecular species  $B$.  

\paragraph{Input-Output relation}
We assume here that the CN has  $N$ different species of input molecules $A_1,\ldots,A_N$. These represent the $N$ input channels and are associated with the   corresponding weights $H_1, \ldots,H_N$. The input is always provided as an exponentially decaying bolus at a particular time  $t_i^s$. Concretely, this means that  time $t=t_i^s$ the CN is brought in contact with a reservoir consisting of $\beta$ (unmodelled) precursor molecules that then decay into $A_i$ molecules with a rate constant $\kappa>0$. A particular consequence of this is that the $A_i$ are not added instantaneously, but will enter the system over a certain time.   This particular procedure is a model choice that has been made for convenience. Different choices are possible and would not impact on the results to be presented. The important point is that there is a sense in which inputs can be provided to particular input channels $i$ by providing a bolus of $A_i$ at a particular time $t$. Finally, note that the CN has a dissipative component. Both $B$ and $H_i$ molecules decay with a fixed rate constant. This enables the system to reach a steady state provided that the input is stationary. \par
The basic idea of the CN is that input boli $A_i$ are converted into internal state molecules $B$. The speed of conversion depends on the amount of weight $H_i$. If at any one time there is enough of $B$ in the system then the learning signal $\mathcal E$ is created by activating $E$ molecules. Once the learning signal is present then some of the $A_i$ are converted into weight molecules, such that the weight of the particular input channel increases. This realises Hebbian learning in the sense that the coincidence of inputs $B$ and output $\mathcal E$ should activate learning following the well known Hebbian tenet ``What fires together, wires together".  
\paragraph{Activation function}
 Precisely what  constitutes ``enough'' $B$ molecules to trigger a learning signal  depends on the precise tuning of the system and will be discussed extensively below.   The link between the internal state molecules $B$ and the learning signal  is often called the  {\em activation function}. More specifically, this  is the  functional dependence of the probability to  observe the activated form $\mathcal{E}$ and  the abundance of $B$. As we will show below, this activation function determines the ability of the CN to learn various kinds of  input biases.  In the simplest, albeit hard to realise case, the activation is a step function. This means that  the learning signal is only generated if  the abundance of $B$ molecules is greater than some threshold $\vartheta$.   
\par
In reality, the activation function will at best approximate a step function. In the CN it is realised  as follows:  In all simulations presented below, we start  at time $t=0$  with  a fixed number of  $E$ molecules. Each of the  $E$ molecule has $m$ binding sites  to which the internal state molecules  $B$ can bind. Once all $m$ binding sites are occupied, then $E$  is converted into its active form $\mathcal E$. We make here the simplifying assumption that the conversion from $E$ to $\mathcal E$ is instantaneous once the last $B$ binds. Similarly, if a $B$ unbinds, then the $\mathcal E$ changes immediately to $E$. In this model, the  balance between $\mathcal E$ and $E$ molecules depends on the binding and unbinding rates of $B$. We assume that there is a cooperative interaction between the $B$ molecules such that unbinding of $B$ from $\mathcal E$ is much slower than unbinding from $E$. With an appropriate choice of rate constants, this system is known to implement ultrasensitivity, i.e. the probability for the fully occupied form of the ligand chain ($\mathcal E$) to exist transitions rapidly from close to 0 to close to 1 as the concentration of ligands approaches a threshold value $\vartheta \approx k_+/k_-$. Such systems are often approximated by the so-called Hill kinetics.  It can be shown that the maximal Hill exponent that can be achieved by such a system is $m$ \citep{Chu_2009}. This means that the chain-length  $m$, which we henceforth shall refer to as the ``nonlinearity'', controls the steepness of the activation function of $\mathcal E$. In the limiting case of $m=\infty$, this will be a step function, whereby the probability to  observe $\mathcal E$ is 0 if  the abundance of $B$  is below a threshold and 1  otherwise. We are limited here to  finite values of $m>0$. In this case, the function is sigmoidal, or a saturating function in the case of $m=1$. The parameter $m$ and hence the steepness of the activation function will turn out to be a crucial factor determining the computational properties of the CN. 
\par
The functional dependence of $\mathcal E$ on $B$ is the analogue of an activation function in SNs. However, note that unlike standard practice in SNs,  the activation function  determines  the probability of a learning signal being present as a function of the abundance of $B$ molecules, i.e. it is a stochastic function.
\paragraph{Learning}
 The weights $H_i$ of a particular input channel $i$ can only be increased if two conditions are fulfilled: \one  the learning signal $\mathcal E$ is present and \two there are still input molecules $A_i$ in the system. In short, learning can only happen if input and output coincide, which is precisely the idea of Hebbian learning. 

\subsection{Associative learning}
\begin{figure}
\includegraphics[width=\textwidth]{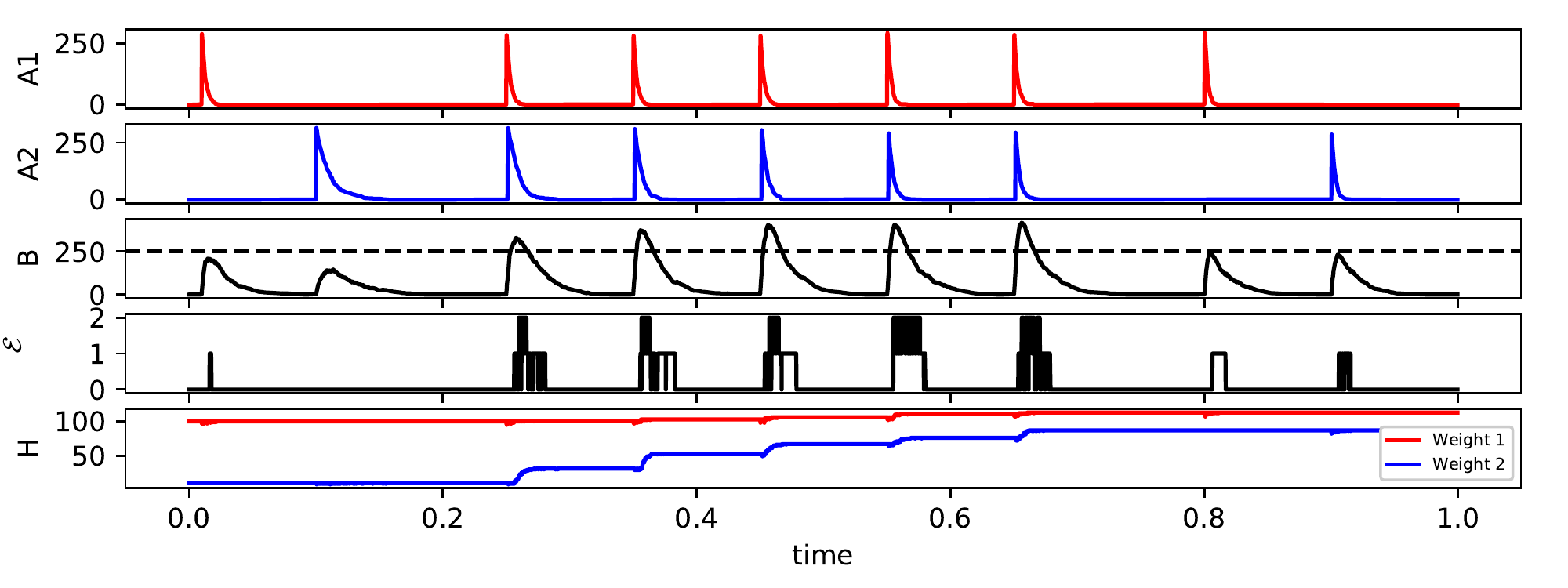}
\caption{
Associative learning in CN.   The first two graphs show the inputs  $A_1$ and $A_2$. Clearly, a single $A_2$ does not lead to a sufficient increase of the internal state $B$, such that no learning signal is triggered. After a few coincidences of $A_1$ and $A_2$  the weights $H_2$ (last graph) have increased sufficiently for $A_2$ to trigger a signal in its own at time $t=0.8$.  Note the increase in weights for the second channel after each coincidence. On the contrary, the weight associated with the first input channel experiences a marginal increase. This shows that the system implements a form of local weight normalisation mechanism similar to the one proposed by \cite{oja_1982}.} 
\label{association_learning}
\end{figure}
%
Throughout this paper we will assume that the dynamics of $A,B$ and $E$ is fast compared to the change in concentration of $H$. This is a crucial assumption to allow the weights to capture long-term properties of inputs; in particular, the weights should not be influenced by high frequency noise present in the system.  Furthermore, we also assume that the lifetime of $\mathcal E$ is short.  For details of the parameters used see table \ref{params_table}. 

We first demonstrate that the  CN  is capable of associative learning; see fig. \ref{association_learning}. To do this, we generate a CN with $N=2$ input channels. Then,  we initialise the CN  with a high  weight for the first channel ($H_1=100$) and a low  weight for the second channel ($H_2=0$).  Furthermore, we set the parameters of the model such that a bolus of $A_1$ is sufficient to trigger an output, but a bolus of $A_2$, corresponding to stimulating the second channel is not; see fig. \ref{association_learning} for details. This also means that presenting simultaneously  both $A_1$ and $A_2$  triggers a learning signal and increases $H_1$ and $H_2$. If  $A_1$ and $A_2$ coincide a few times, then the weights of $A_2$  have increased sufficiently so that a bolus of $A_2$ can push the internal state of the system over the threshold on its own; see fig. \ref{association_learning}. This demonstrates associative learning. Note that unlike some previous molecular implementations of associative learning (e.g. \citep{Fernando2009}), the system  requires several coincidences before it learns the association. It is thus robust against noise. It will also  readily  unlearn the correlation if input patterns change. 

\subsection{Full Hebbian learning}

We now show that the ability of  the CN to learn extends to full Hebbian learning with an arbitrary number of $N$ input channels.  First we consider the FB task. To do this, we  provide random boli  to each of the $N$ input channels.  Random here means that the waiting time between two successive boli of  $A_i$ is distributed  according to an exponential distribution with parameter $1/f_i$, where $f_i$ is the frequency of the input boli to channel $i$.  The  CN should then  detect the difference in frequencies $f_i$ between input channels. We consider the FB task as solved if (after a transient period) the ordering of the abundances of weights reflects the input frequencies, i.e. the number of $H_i$ should be higher than the number of $H_j$ if $f_i> f_j$. 
\par
\begin{figure}
\includegraphics[width=\textwidth]{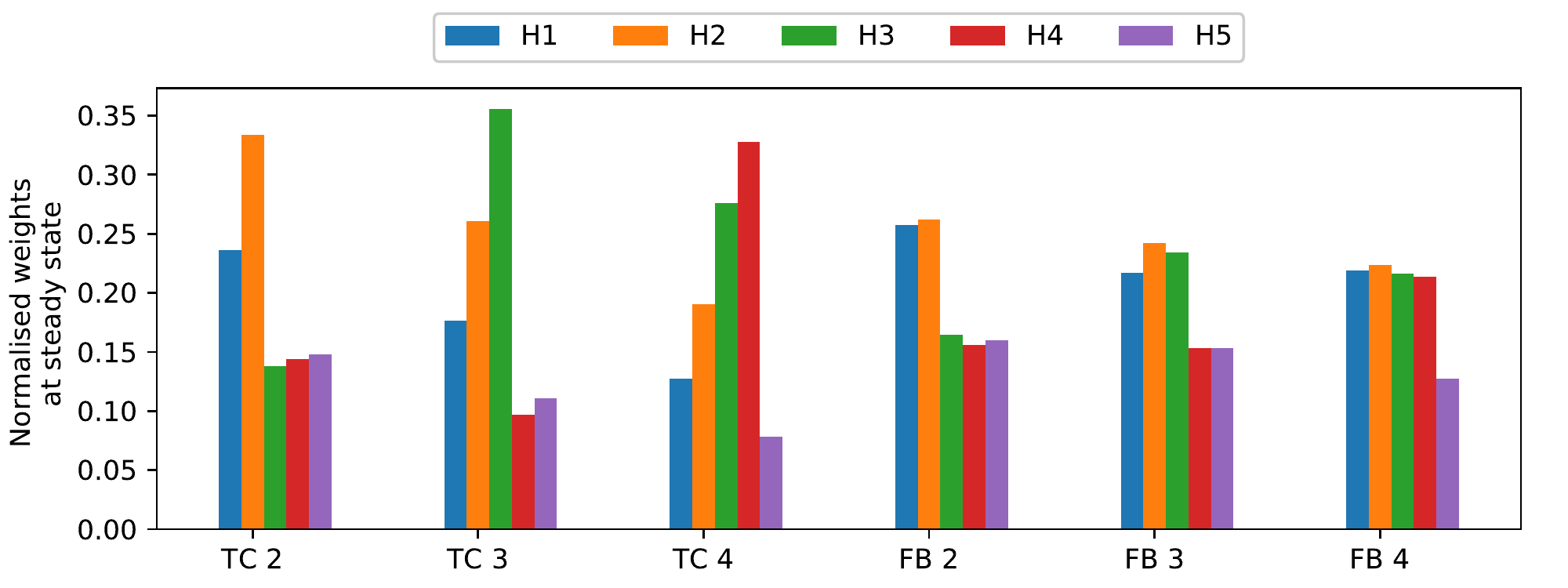}
\caption{Normalised weights for a variety of TC and FB tasks. The first (blue) bar refers to the first weight, the second (orange) to the weight for the second channel and so on.     Each value represents the average over 300 time units of a single simulation.  Data was only collected after the weights reached the steady state (after 700 time units).  In all experiments we set the number of $E_1$ molecules at the start of the simulation to 40. The nonlinearity was set  to $m=5$ for the TC, and $m=1$ for FB.}
\label{tasks_fig}
\end{figure}

In order to test this, we consider 3 variants of the FB task: We use  a CN with $N=5$ and $m=1$ and assume that boli to the first two channels input  come at a frequency of 4Hz whereas  channels 3, 4 and 5  fire at a  frequency of  2 Hz; we call this variant FB 2. Similarly, for FB 3 and FB 4 the first 3 and 4 channels respectively fire at the high frequency.  Fig.  \ref{tasks_fig} shows the steady state weights for  each of the three tasks. As expected, in each of the experiments the weights of the high-frequency inputs are higher when compared to the low frequency inputs. We conclude, that the CN can work as a frequency detector.
\par
A more interesting scenario is the direct generalisation of the associative learning task to an arbitrary number of input channels.   For this problem we assume that all input frequencies are the same, i.e. $f_i=f_j$ for all $i,j\leq N$. Instead of differences in frequency, we  allow temporal correlations between input boli of some channels.  This means that there are pairs of channels $i,j$ and time windows $\Delta \tau$ such that the probability to observe a bolus of $j$ between time $t_0$ and  $t_0 + \Delta\tau$  --- where $t_0$ is the time where input $A_i$ fired --- is greater than  the probability to observe a bolus  between $t_0-\Delta \tau$ and time $t_0$. In practice, we implement such correlations  as follows: If $A_1$ and $A_2$ are temporally correlated then  each  bolus of   $A_1$ is  followed by  a bolus of  $A_2$ after a time  period of  $\delta  + \xi$;  with $\delta$ being a fixed number and $\xi$ is a  random variable drawn from normal distribution  with $\mu=0$ and $\sigma^2=0.0001$ for each bolus. In all simulations we present below, the input frequency of all channels is set to 2Hz. 
\par
The CN can solve the TC task in the sense that after a transient period  the weights indicate which channels are correlated and the temporal order implied by the correlation, i.e. if $A_i$ tends to precede $A_j$, then  the abundance of weight  $H_i$ should be higher than the abundance of $H_j$. Furthermore, if $A_i$ is correlated with some other channel $k$, but $A_j$ is not, then the abundance of  $H_i$ must be greater than that of $H_j$. 
In order to test the ability of the system to detect TC biases, we  initialised the CN with $N=5$ input channels. Initially the weight molecules $H_i=0$.  We then determined the steady state weights. We compared four different  scenarios: \one there are correlations between $A_1$ and $A_2$ (TC 2), \two  $A_1,A_2, A_3$ (TC 3), and \three  $A_1,A_2,A_3, A_4$ (TC 4). The temporal order is always in ascending order of the index, such that in the last example, $A_1$ occurs before $A_2$, which in turn occurs before $A_3$.  We find that the behaviour of the CN is as expected; see fig. \ref{tasks_fig}. At steady states the weights reflect the correlation between input channels, including the temporal ordering.

\subsection{Correlation detection}
We now examine how the nonlinearity parameter $m$ impacts on the ability of the system to compute.  We consider two extreme cases: Firstly, the case of minimal non-linearity (i.e.  $m=1$) and the secondly, the limiting case of maximal nonlinearity (i.e. $m=\infty$). This latter case would correspond to an activation function that is a step function. Strictly speaking, the CN cannot realise a pure step function, but it still provides valuable insight to think through this limiting case.
\par
 We consider first this latter scenario with a CN with two inputs $A_1$ and $A_2$. In this case, there will be a learning signal $\mathcal E$ in the CN if the abundance of $B$ crosses the threshold $\vartheta$. Let us now assume that the parameters are set such that a single bolus of either $A_1$ or $A_2$ is not sufficient to push the abundance of $B$ over the threshold, but a coincidence of both is:  
\begin{itemize}
\item
A single bolus of $A_1$ will not lead to a threshold crossing. No learning signal is generated and  weights are not increased.
\item
If a bolus of $A_1$ coincides with a bolus of $A_2$ then this may lead to a crossing of the threshold of the internal state. A learning signal is generated. Weights for both input channels 1 and 2 are increased (although typically not by equal amounts).   
\end{itemize} 
Next consider an activation function tuned to the opposite extreme, i.e. $m=1$. It will still be true that both $A_1$ and $A_2$ are required to push the abundance of $B$ across the threshold. However, the learning behaviour of the CN will be different:
\begin{itemize}
\item
A single bolus of $A_1$ will not lead to a threshold crossing. A learning signal may still be generated even below the threshold because the activation function is not a strict step function.  The weight $H_1$ will increase by some amount, depending on the bolus size. 
\item
If a bolus of $A_1$ coincides with a bolus of $A_2$ then this will lead to more learning signal being  generated than in the case of  $A_1$ only.  As a result,  the  weights for both input channel 1 and 2 are increased by more than if they had occurred separately.   
\end{itemize} 
\par
These two extreme cases illustrate how the CN integrates over input. In the case of low non-linearity the weights of a channel will be a weighted sum over all input events of this channel. The weights will be  higher  for  channels  whose boli coincide often.   On the other hand, a step-like activation function will integrate only over those events where the threshold was crossed, thus specifically detect coincidences.  From this we can derive two conjectures:
\begin{itemize}
\item
The higher the nonlinearity, the better the CN at detecting coincidences. Low non-linearity still allows coincidence detection, but in a much weaker form.
\item
As the bolus size increases, the CN will lose its ability to detect coincidences, especially when the bolus size is so large that a single bolus is sufficient to push the abundance of $B$ over the threshold. In this case, a single input spike can saturate the activation function, thus undermining the ability of the system to detect coincidences effectively. 
\end{itemize} 
\par
\begin{figure}
\centering
\includegraphics[width=\textwidth]{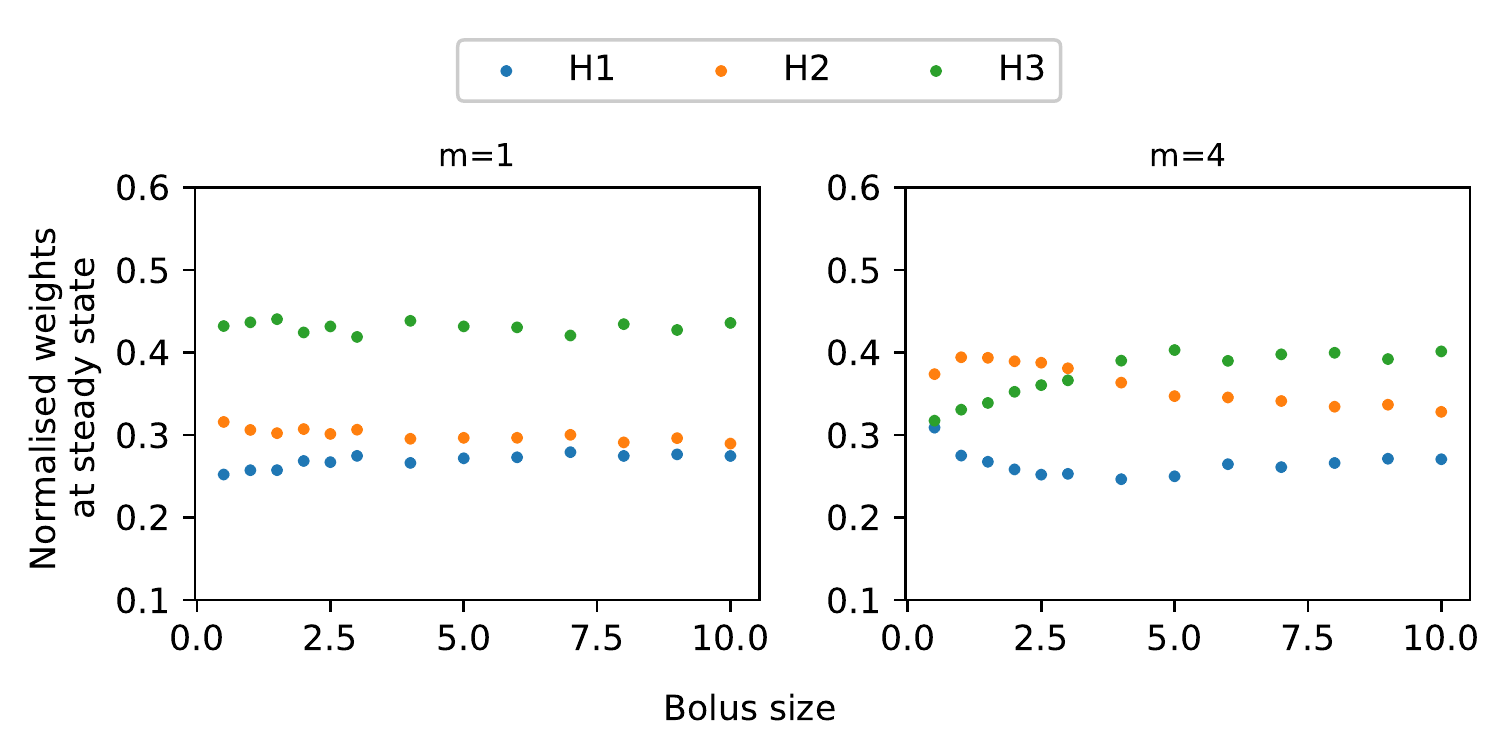}
\caption{The weights as a function of bolus size for a CN with 3 inputs. Input $A_3$ (green) is provided at 4Hz, $A_1$ and $A_2$ are correlated with a $\delta= 0.0047 $ but they are only provided at 2Hz. The graph shows the normalised weights at steady state corresponding to the input channels for different boli-sizes (here reported as a fraction of the threshold). From left to right, the non-linearity increases. For $m=2$ the system detects the higher frequency of  $A_1$ as indicated by its high weight throughout. The weight of $A_2$ (orange) is only slightly higher than the weight of $A_3$, indicating that the CN detects the coincidence only to some  limited extent. For increased non-linearity, the CN indicates the correlation because weights are only increased when coincidences occur. If the bolus size is increased, then a single bolus is sufficient to cross the threshold and the frequency bias is recognised again.  }
\label{meanH_bolus_m1_5_mixHz}
\end{figure}

In order to check these conjectures, we simulated a CN with 3 inputs, where $A_1$ and $A_2$ are correlated and $A_3$ fires at twice the frequency of $A_1$ and $A_2$. We considered the minimally nonlinear case ($m=1$) and a moderate non-linearity ($m=4$); Fig. \ref{meanH_bolus_m1_5_mixHz} shows the weights as a function of the bolus size. The minimal non-linear CN detects both coincidences and frequency differences, but loses its ability to detect coincidences as the bolus size increases. This is consistent with the above formulated hypothesis. In contrast, for the non-linear CN and moderately low bolus-sizes the weights indicate the coincidences strongly (i.e. the weights $H_2$ are highest). Yet,  as the bolus size increases the non-linear CN also loses its ability to detect coincidences.
\par
\begin{figure}
\includegraphics[width=0.5\textwidth]{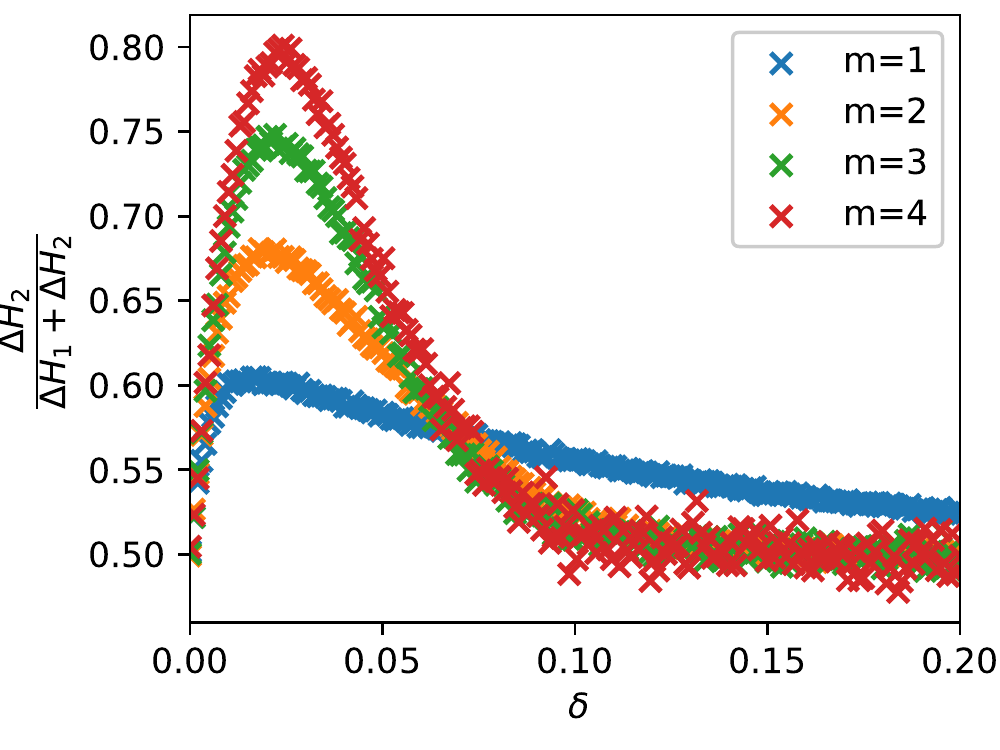}
\includegraphics[width=0.5\textwidth]{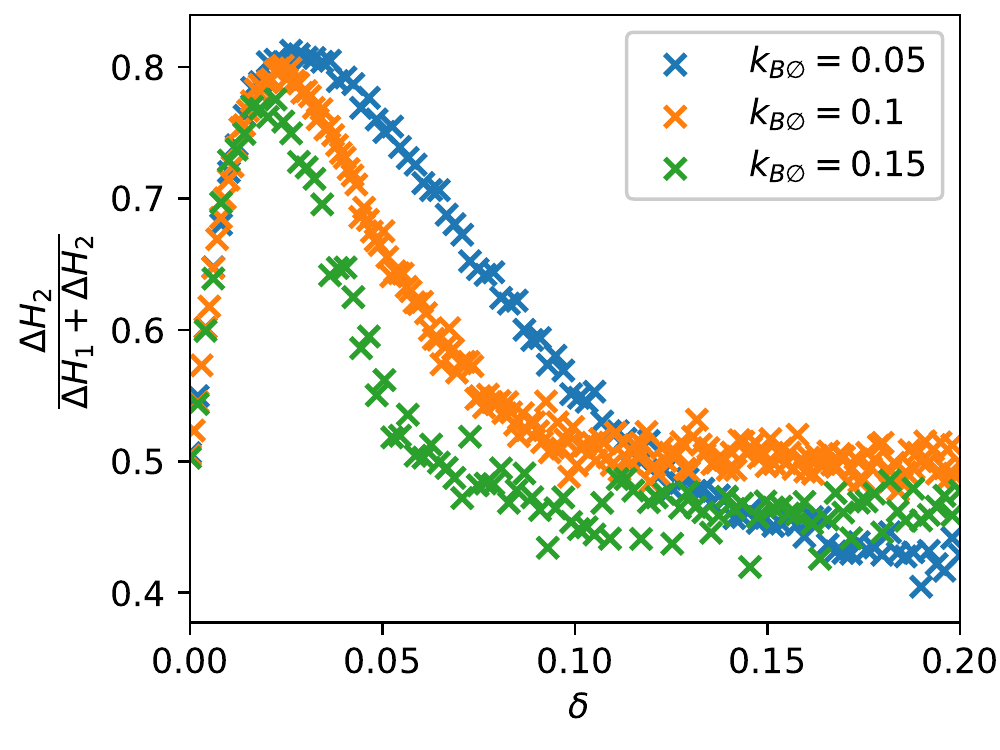}\label{long_ratio_sum} 
\caption{Differential weight increase for different chain lengths. A value of 1 means that only the first input channel received weight accumulation. A value of $0.5$ means that the weights of both channels were updated equally (left).  Same, but for different removal rates of $B$ (right). The faster the removal, the more specific the coincidence detection, i.e. inputs need to occur within a narrower window.  For both graph points were computed as follows:  We simulated a CN with two input channels only and an initial condition  of $H_1, H_2=0$. At time $t=0$ we provided a bolus of $A_1$ and after a variable time we provided the bolus $A_2$. We then continued the simulation for another $0.2$ time units. The $y$-axis records the relative increase of $H_2$ averaged over 1000 repetitions.  }
\label{dist_m_long}
\end{figure}

Next, we check how the coincidence detection depends on the time-delay between the correlated signals. To do this we created a scenario where we provided two boli to the system. The first bolus $A_1$ comes at a fixed time and the second one a fixed time period $\delta$ thereafter.  We then vary the length of $\delta$ and  record the accumulation of weights $H_1$ as a fraction of total weight accumulation. Fig. \ref{dist_m_long} confirms that the CN with the low non-linearity is less sensitive to short coincidences, but can detect coincidences over a longer period of time. In contrast, as the non-linearity increases, the differential weight update becomes more specific, but also  more limited in  its ability to detect coincidences that are far apart. In the particular case, for a $\delta> 0.1$ the CN with $m>1$ does not detect any coincidences any more, whereas the case of $m=1$ shows some differential weight update throughout.   
\par
\begin{figure}
\centering
\includegraphics[width=\textwidth]{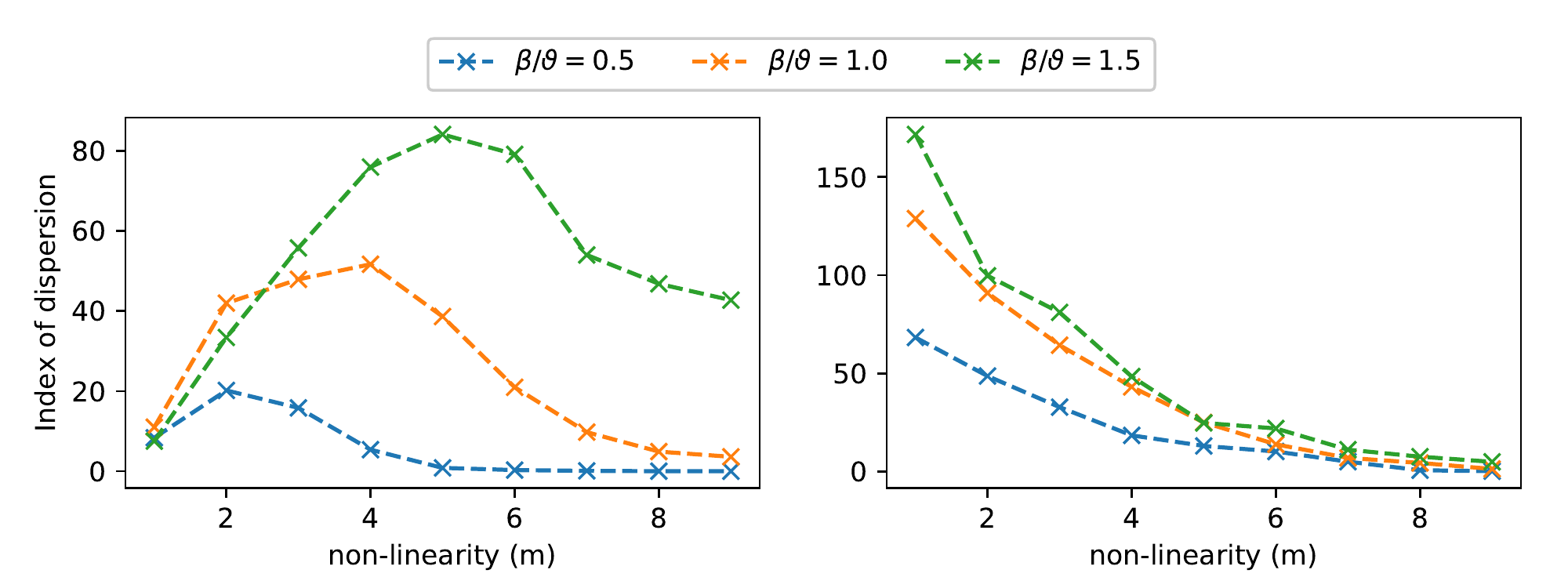}
\caption{Index of dispersion for different bolus sizes $\beta$ expressed as a fraction of the threshold $\vartheta$. We show that for TC 2 (left) and FB 2 (right) tasks. The index of dispersion measures how different the steady state weights are from one another, and hence indicates how well the CN distinguished between input channels. Completely unbiased input would give an index of dispersion of $\approx 0$. The graph shows that for the TC task, there is an optimal nonlinearity. Increasing the bolus size, increases the optimal non-linearity, which is consistent with the fact that the optimum is due to resource starvation. }
\label{iod_volume}
\end{figure}
%
%
Next, we tested the conjecture that the TC can be solved more effectively by the CN when the nonlinearity is higher.  To do this, we generated a CN with $N=5$ input channels on the TC 2 task. We then trained the CN for nonlinearities $m=1,\dots, 10$. As a measure of the ability of the system to distinguish the weights, we used the index of dispersion, i.e. the standard deviation divided by the mean of the weights.  A higher index of dispersion indicates more heterogeneity of the weights and hence a better ability of the system to discriminate between the different frequencies.  Consistently with  our hypothesis we found that the ability to discriminate between frequencies increases with the nonlinearity. However, it does so only up to a point (the optimal nonlinearity), beyond which the index of dispersion reduces again; see fig. \ref{iod_volume}.  Increasing the bolus size, i.e. increasing the number of $A_i$ that are contained within a single bolus,  shifts the optimal nonlinearity to the right.  This suggests that the decline in the performance of the CN for higher chain lengths is due to a resource starvation. The realisation of the sigmoidal function, i.e. the thresholding reactions in table \ref{reaction_table}, withdraws $m$ molecules of $B$  from the system. As a consequence, the CN is no longer able to represent its internal state efficiently and the activation function is distorted. If the total abundance of $B$ is high compared to $E$, then this effect is negligible. We interpret this as a resource cost of computing non-linearity. The higher $m$, the higher the bolus size required to faithfully implement the activation function. 
\subsection{Frequency detection}

We have established that the CN is a coincidence detector, which is why it is able to solve the TC task.  We showed above that the CN can also perform the  FB problem, i.e. it can record frequencies of input signals. The FB task is fundamentally about integrating over input, which  can be done naturally in chemical systems. Indeed  it can be done by systems that are much simpler than the CN, for example:  \ce{$A_{i}$ ->[d] $\varnothing$}.  For appropriately chosen values of $d$, the steady state value of $A_i$ would then reflect the input frequency. To understand this, note that the input frequency determines the  rate of increase of $A_i$. This rate, divided by the decay rate constant $d$ then determines the steady state abundance of $A_i$, such that $A_i$ trivially records its own frequency. This system is the minimal and ideal frequency detector.
\par
The CN itself is not an ideal frequency detector because all weight updates are mediated by the internal state $B$. Hence, the weights are always convolutions over all inputs. The weights thus reflect both frequency bias and temporal correlations. In many applications this may be desired, but  sometimes it may not be. We now consider the conditions necessary to turn the CN into a pure frequency detector, i.e  a system that indicates only FB, but not TC. One possibility is to set the parameters such that the CN approximates the minimal system. This could be   achieved by  setting  $k_\mathrm{BA} \ll  k_\mathrm{AB}$ and all other rate constants very high in comparison to $k_\mathrm{AB}$.   The second possibility is  to tune the CN such that  a single bolus saturates the threshold. In this case, the strength of the learning signal does not depend on the number of boli that are active at any one time. A single bolus will trigger the maximal learning signal. This is confirmed by fig. \ref{meanH_bolus_m1_5_mixHz}, which shows that as the bolus size increases, the system becomes increasingly unable to detect temporal correlations, but remains sensitive to frequency differences. 
%
%

\subsection{A biological model}

The basic model of the  CN, as presented in table \ref{reaction_table} is thermodynamically plausible and has the benefit of being easy to simulate  and analyse. However, it is biologically implausible. As written in table \ref{reaction_table} the molecular species $A_i, H_i$ and $B$ would have to be interpreted as conformations of the same molecule with  different  energy levels. Additionally, we require that these different conformations have specific enzymatic properties. Molecules with the required properties are not known currently, and it is unlikely that they will be discovered or engineered in the near future.  
\par
\begin{figure}
\centering
\includegraphics[width=0.67\textwidth]{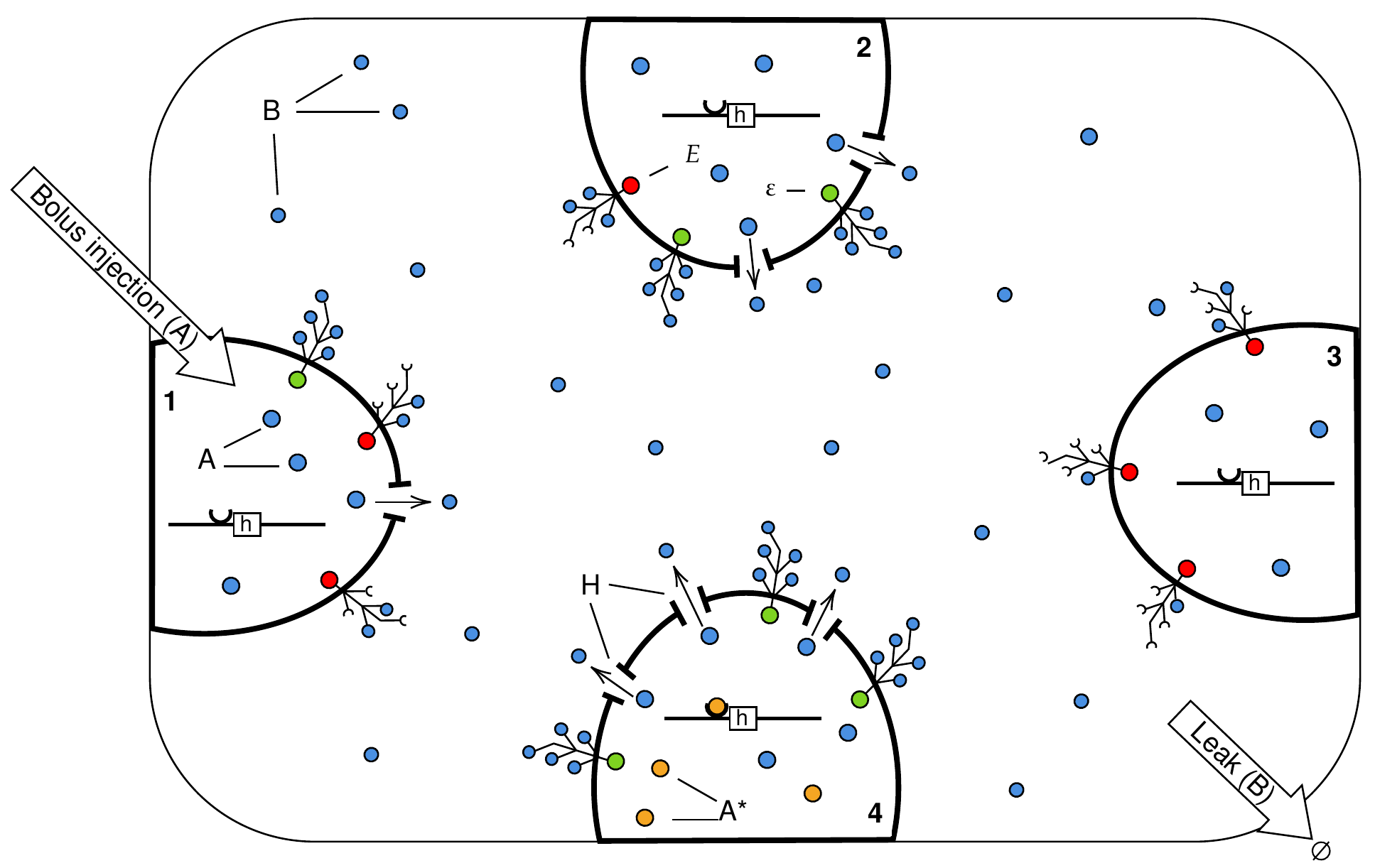}
 \caption{Graphical representation of a compartment-version of the CN. $A_i$ and $A_j$ are the same molecular species but contained in different  compartments $i$ and $j$ respectively. We allow for an activated form of $A$, denoted by $A^*$, which binds to the promotor site of $h$ and activates its expression.   $H$ is an active transporter molecules for $A$. The internal state molecule $B$ is any of the $A_i$ when in the outermost compartment. We assume that each compartment has a trans-membrane protein $E$ with $m$ extra-cellular binding sites. If all $m$ binding sites are occupied by $B$, then the internal site becomes active (indicated by green) and can catalyse the activation of $A$. }
 \label{biomodel}
 \end{figure}

As we will show now it is possible to re-interpret the basic model of the CN (table \ref{reaction_table}) so as to get a model whose elements are easily recognisable by computational biologists as common mechanistic motifs. This requires only relatively minor adjustments of the reactions themselves, but a fundamental re-interpretation of what the reactions mean. For the list of  modified reactions see table \ref{reaction_table2} and fig. \ref{biomodel} for a graphical illustration of the intuition behind the model.
\par
 The main difference  between the basic CN and the biological version is that the latter is compartmentalised. While in the basic model the index (as in $A_i$ and $H_i$) referred to different species that exist in the same volume, it should now be interpreted as indicating different compartments that separate the same molecular species.  These compartments, which  are themselves enveloped in a further compartment (the ``extra-cellular space''),  could be thought of as  individual  bacterial cells or else artificially generated membranes with a minimal genome.
Input is provided by boli of the molecular species $A$ into the compartment $i$. There is now also an activated form of $A$, denoted by $A^*$.  The conversion from $A$ to $A^*$ is catalysed by the learning signal $\mathcal E$.  Also new is that each compartment contains a  gene  $h$ that codes for the molecule $H$ (we suppress the index indicating the compartment). Expression of the gene is activated by $A^*$ binding to the promoter site of $h$. We also allow a low leak expression by the  unactivated gene (denoted as $h_0$ in table \ref{reaction_table2}). Gene activation of this type is frequently modelled using Michaelis-Menten kinetics, thus reproducing in good approximation the corresponding enzyme kinetics in the basic model of the CN.
   The molecules of type $H$ are now transporters for $A$.  We then interpret the  conversion of $A_i$ to $B$ as export of $A$ from compartment $i$ to the extra-cellular space. In this interpretation the molecular species  $B$ is then the same as   $A$ but  contained directly in the outer compartment. The rate of export of $A$ is specific to each compartment in that it depends on the abundance of $H$ in this compartment.    Finally, we interpret the $E$ molecules as transmembrane proteins that are embedded in the membrane of each compartment. Their extra-cellular site  has $m$ binding sites for $B$ molecules which bind cooperatively. When all sites are occupied then the intra-cellular part is activated, i.e. becomes $\mathcal E$. In its activated form it can convert $A$ to $A^*$.
\par
Another  difference between the two versions of the models is that the  molecule $E$ is now specific to each membrane. The minimum number of copies of $E$ is thus $N$ whereas in the basic model a single copy of $E$ at time $t=0$ could be sufficient.  This has two consequences. Firstly, at any particular time the number of occupied binding sites will typically  be different across the different $N$ compartments. This is a source of additional variability.   Moreover, since the number of copies of $E$ is higher than in the basic CN, the model in table \ref{reaction_table2}  is more susceptible to starvation of $B$ as a result of the extra-cellular binding sites withdrawing molecules from the outer compartment.  Both of these potential problems can be overcome by tuning the model such that the abundance of $B$ molecules is high in comparison to $E$ molecules.  
\par
\begin{figure}
\centering
\includegraphics[width=\textwidth]{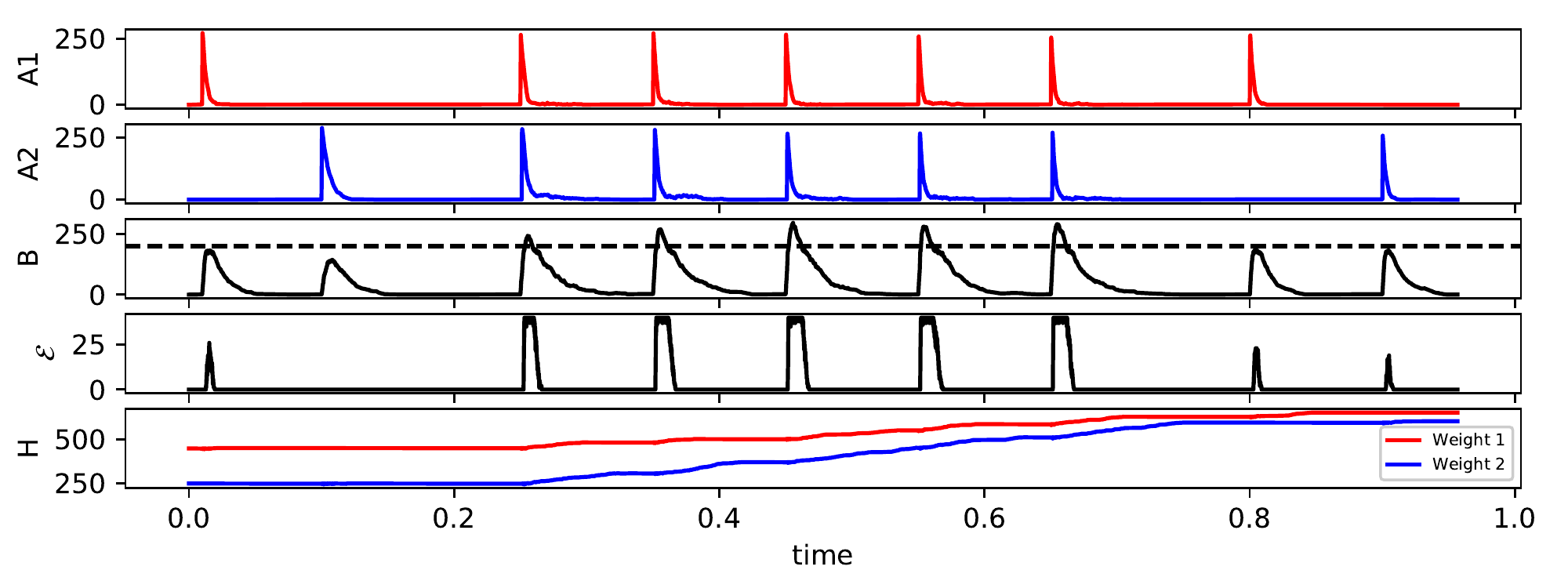} \\
\includegraphics[width=\textwidth]{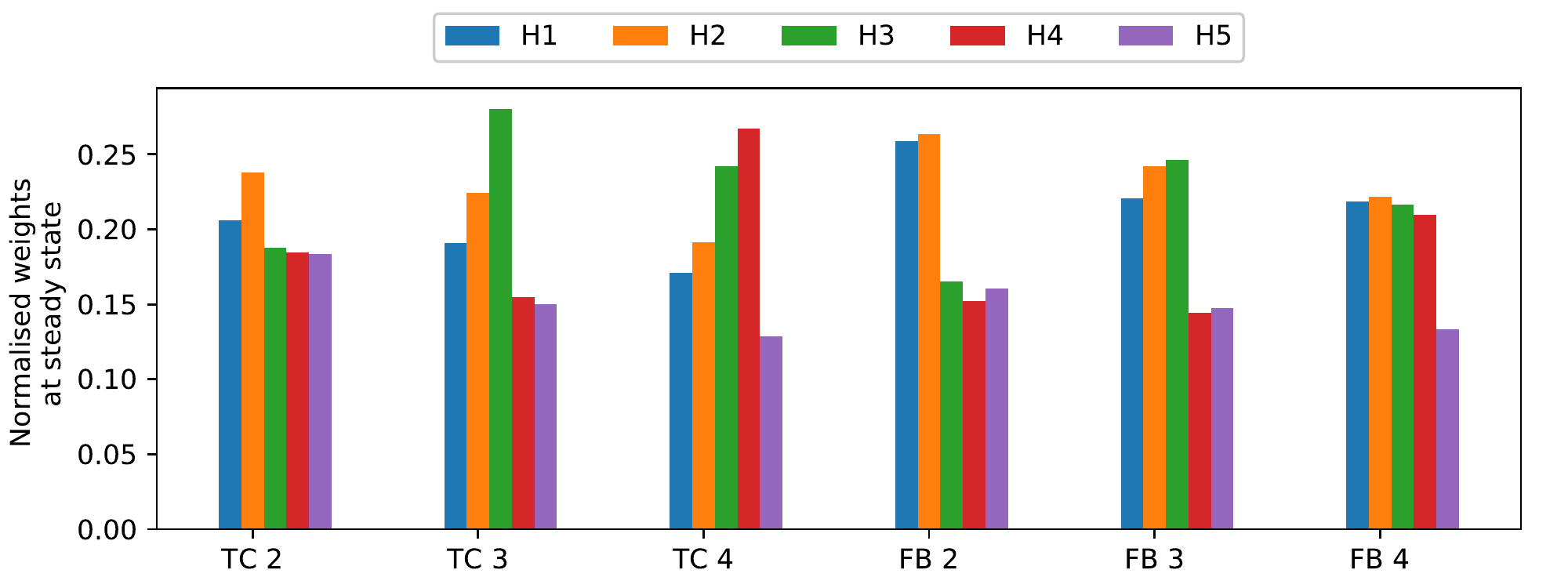}
 \caption{Associative (top) and  Hebbian learning (bottom) for the biologically plausible version of the CN. The experimental setup is as in fig. \ref{association_learning}, and fig. \ref{tasks_fig} respectively. For the parameters used see table \ref{params_table2}. Both experiments approximated the ligand dynamics by a Hill function, in order to speed up the simulations.}
 \label{biomodel_res}
 \end{figure}

This highlights that the difference between the basic CN and its biological interpretation are deeper than the list of reaction suggests. We checked that the biological interpretation still allows the same phenomenological behaviour, provided that the parameters are adjusted. Fig. \ref{biomodel_res} confirms that both associative learning and full Hebbian learning (with 5 input channels) is possible with this model. 

\begin{table}
\centering
\begin{tabular}{|l|l|}
\hline
Input & \ce{I_n <=>[k_{IA}][k_{AI}] A }  \\ \cline{2-2}
\hline
\multirow{2}{*}{Activation function}
& \ce{B + E_{i} <=>[k^{+}][k^{-}] E_{i+1}}, \qquad $i< m-1$ \\ \cline{2-2}
& \ce{B + E_{m-1} <=>[k^{+}][k^{-}_{last}] $\mathcal{E}$}   \\
\hline
\multirow{10}{*}{Learning}
& \ce{$\mathcal{E}$  + A <=>[k_{AE}][k_{EA}] $\mathcal{E}$A <=>[k_{A*E}][k_{EA*}] $\mathcal{E}$ + A^*} \\  \cline{2-2}
& \ce{A^* <=>[k_{A*A}][k_{AA*}] A } \\  \cline{2-2}
& \ce{$h_0$ + $A^*$   <=>[A*h][hA*]  h }  \\\cline{2-2}
& \ce{$h_0$  ->[$k_\mathrm{leak}$] H_n + h_0 }  \\ \cline{2-2}
& \ce{$h$  ->[$k_{h}$] H_n + h }  \\ \cline{2-2}
& \ce{A + H <=>[k_{AH}][k_{HA}] AH <=>[k_{HB}][k_{BH}] B + H}   \\
\hline
\multirow{2}{*}{Leak}
& \ce{H ->[k_{H$\varnothing$}] $\varnothing$}   \\ \cline{2-2}
& \ce{B ->[k_{B$\varnothing$}] $\varnothing$}   \\
\hline
\end{tabular}
\caption{List of chemical reactions in a single CN unit interpreted as a cell. Molecular species $A,E,\mathcal E, h_0,h$ and $H$ are compartmentalised. Each compartment has a gene $h_0$ which when activated by $A^*$ can express a transporter $H$.}
\label{reaction_table2}
\end{table}

\section{Discussion}

To our knowledge the CN presented here is the first  fully autonomous design for a chemical systems capable of full  Hebbian learning. The model is, at least in principle, fully scalable.  Previous attempts (e.g. \citep{mcgregor_cn,Fernando2009}) were limited to 2 input associative learning or they were not fully autonomous (e.g. \citep{chemperceptron2}).  Our proposed basic model is not biologically realistic, but it is thermodynamically consistent and could be analysed with respect to its minimal energy dissipation. The second design we proposed is biologically more realistic. While the second model is not directly translatable into a synthetic biology design, its basic building blocks (i.e. activated gene expression, export of molecules, cooperative binding) are  recognisable biological components.  
\par
The CN is  closely analogous to the  {\em leaky integrate and fire} (LIF) neuron which is a commonly used    continuous time SN \citep{Fil_2020, gerstner_kistler_2002}.  The LIF neuron is a powerful computational unit \citep{maass1997, Maass1998, Gutig2014, Gutig2006}. The CN inherits this.
\par
One of the attractive features of the (basic) CN neuron is that it is a micro-reversible model and as such it is a thermodynamically plausible model. While a thorough analysis of the energy requirements of the system is beyond the   scope of this article, we note that the physical plausibility of the model has highlighted resource requirements of the computation.  In particular, we found that increasing the non-linearity comes at an additional cost in resource. The CN suffers from starvation of $B$ molecules as the chain-length of realising the activation function increases.  For a sufficiently high number of $m$, this leads to a breakdown of the mechanisms and the system loses its ability to detect coincidences, as illustrated  in fig. \ref{iod_volume}.  The CN is better at solving the TC task when the nonlinearity is higher. At the same time, the non-linearity  requires resources that cannot be met any more by the system. This ``starvation'' effect can be alleviated by increasing the bolus size (while keeping the threshold fixed); see fig. \ref{iod_volume}.   The minimal amount of work required to add particles to the CN is proportional to the number of particles, i.e. the bolus size. Less visibly, there is also an additional expenditure of energy due to the fact that the $B$ and $H$ molecules need to be removed from the system, which again requires a concentration gradient to be maintained. Hence, at least in this model the non-linearity comes directly at a thermodynamical cost. 
To the best of our knowledge, it is an open question whether the computation of non-linearities necessarily comes at a higher thermodynamic cost or whether this is just a side effect of the computational medium used here, i.e. chemical reactions. 
\bibliographystyle{plainnat}
\bibliography{biblio} 
\appendix
\section{CN chemical reaction network details}
\label{Appendix1}
\begin{table}[H]
\centering
\begin{tabular}{|l|l|} 
\hline
\multirow{2}{*}{Input}              
& $k_{IA}=10$, $k_{AI}=0.000001$  \\ \cline{2-2}
& $k_{AB}=0.1$, $k_{BA}=0.000001$  \\ 
\hline
\multirow{2}{*}{Activation function }
& $k^{+}=1$, $k^{-}=5$ \\ \cline{2-2}
& $k^{-}_{last}=0.5$  \\ 
\hline
\multirow{2}{*}{Learning}
& $k_{AE}=0.05$, $k_{EA}=0.000001$, $k_{EH}=100$, $k_{HE}=0.000001$   \\
& $k_{AH}=0.001$, $k_{HA}=0.000001$, $k_{HB}=100$, $k_{BH}=0.000001$ \\ 
\hline
\multirow{2}{*}{Leak}
& $k_{H\varnothing}=0.0003$  \\ \cline{2-2}
& $k_{B\varnothing}=0.1$ \\ 
\hline
\end{tabular}
\caption{List of reaction rate constants in the CN model.}
\label{params_table}
\end{table}
\begin{table}[H]
\centering
\begin{tabular}{|l|l|}
\hline
Bolus & $k_{IA}=10$, $k_{AI}=0.000001$  \\ 
\hline 
Activation function (Hill) & $h = 10$ \\
\hline
\multirow{6}{*}{Learning}
& $k_{AE}=0.2$, $k_{EA}=0.000001$, $k_{A*E}=0.2$, $k_{EA*}=0.000001$,\\ \cline{2-2}
& $k_{A*A}=0.05$, $k_{AA*}=0.000001$ \\ \cline{2-2}
& $k_{A*h}=1$, $k_{hA*}=0.1$ \\ \cline{2-2}
& $k_{leak}=0.0001$ \\
& $k_{h}=1$ \\
&  $k_{AH}=0.03$, $k_{HA}=0.000001$, $k_{HB}=100$, $k_{BH}=0.000001$  \\
\hline
\multirow{2}{*}{Leak}
& $k_{H\varnothing}=0.0003$   \\ \cline{2-2}
& $k_{B\varnothing}=0.1$   \\
\hline
\end{tabular}
\caption{List of reaction rate constants in a biological interpretation of the CN model.}
\label{params_table2}
\end{table}
%
%
%
%
\end{document}